%% file: root.tex
\documentclass[pdflatex,sn-mathphys-num]{sn-jnl}

\usepackage{graphicx}
\usepackage{multirow}
\usepackage{amsmath,amssymb,amsfonts}
\usepackage{amsthm}
\usepackage{mathrsfs}
\usepackage[title]{appendix}
\usepackage{xcolor}
\usepackage{textcomp}
\usepackage{manyfoot}
\usepackage{booktabs}
\usepackage{algorithm}
\usepackage{algorithmicx}
\usepackage{algpseudocode}
\usepackage{listings}
\usepackage{float}

\theoremstyle{thmstyleone}

\theoremstyle{thmstyletwo}

\theoremstyle{thmstylethree}

\begin{document}

\title[Article Title]{Multi-View Camera Perception System for Variant-Aware Autonomous Vehicle Inspection and Defect
Detection}



\author*[1]{\fnm{Yash} \sur{Kulkarni}}\email{yaskul31@gmail.com}
\author[2]{\fnm{Raman} \sur{Jha}}
\author[3]{\fnm{Renu} \sur{Kachhoria}}

\affil[1]{\orgdiv{Department of Computer Science}, \orgname{Vishwakarma Institute of Information Technology},
\orgaddress{\city{Pune}, \country{India}}}

\affil[2]{\orgdiv{Department of Electrical and Computer Engineering}, \orgname{New York University},
\orgaddress{\city{New York}, \country{USA}}}

\affil[3]{\orgdiv{Department of Computer Science}, \orgname{Vishwakarma Institute of Technology},
\orgaddress{\city{Pune}, \country{India}}}




\abstract{Ensuring that every vehicle leaving a modern production line is built to the correct \emph{variant} specification and is free from visible defects is an increasingly complex challenge. We present the \textbf{Automated Vehicle Inspection (AVI)} platform, an end-to-end, \emph{multi-view} perception system that couples deep-learning detectors with a semantic rule engine to deliver \emph{variant-aware} quality control in real time.  
Eleven synchronized cameras capture a full 360° sweep of each vehicle; task-specific views are then routed to specialised modules: YOLOv8 for part detection, EfficientNet for ICE/EV classification, Gemini-1.5 Flash for mascot OCR, and YOLOv8-Seg for scratch-and-dent segmentation.  A view-aware fusion layer standardises evidence, while a VIN-conditioned rule engine compares detected features against the expected manifest, producing an interpretable pass/fail report in \(\approx\! 300\,\text{ms}\).  
On a mixed data set of Original Equipment Manufacturer(OEM) vehicle data sets of four distinct models plus public scratch/dent images, AVI achieves \textbf{ 93 \%} verification accuracy, \textbf{86 \%} defect-detection recall, and sustains \(\mathbf{3.3}\) vehicles/min, surpassing single-view or no segmentation baselines by large margins. To our knowledge, this is the first publicly reported system that unifies multi-camera feature validation with defect detection in a deployable automotive setting in industry.}

\keywords{Semantic occupancy prediction, Multi-sensor data fusion, Vehicle Inspection, Defect Detection, Quality Control, Scene understanding, Autonomous driving, Feature fusion }

\maketitle

\input{01_introduction}
\input{02_related_work}
\input{03_method}

\input{04_experiment}

\input{05_discussion}
\input{06_conclusion}

\section{Declarations}
\begin{itemize}

\item \textbf{Acknowledgements}\\
We would like to thank Vishwakarma Institute of Technology and New York University for the guidance and support. The authors also wish to thank the reviewers for their valuable and thoughtful comments, which helped enhance the overall quality of the manuscript.\\

\item \textbf{Funding}\\
No Funding.\\

\item \textbf{Author contribution\\}
Yash Kulkarni conducted the model development and experiments and led the overall project. Raman Jha carried out the literature review, data processing, and article writing. Renu Kachhoria proofread the manuscript and provided constructive feedback. All authors read and approved the final version of the manuscript.\\

\item \textbf{Corresponding author:} \\
Correspondence to \href{Yash Kulkarni}{yaskul31@gmail.com}. \\

\item \textbf{Conflict of interest/Competing interests} \\
The authors confirm that they have no conflicts of interest with any third-party organizations related to this work.\\

\item \textbf{Data availability material (data transparency)}   \\
The Vehicle damage data is publicly available at: https://universe.roboflow.com/axion-technical-service-pvt-ltd/damage-severity-bwiya.     
The vehicle parts data is publicly available at: https://universe.roboflow.com/habibullah-hmpb8/car-parts-chf9t. \\

\item \textbf{Code availability (software application or custom code)}\\
Not applicable 

\end{itemize}

\bibliography{root}

\end{document}

%% file: 01_introduction.tex
\section{Introduction}
\label{sec:intro}
Ensuring that every vehicle leaving a modern production line matches its intended \emph{variant} and is free of visible defects is a central requirement of Industry~4.0 manufacturing, where autonomy, connectivity, and data-driven QA are becoming standard practice~\cite{cronin2019state}. As customization proliferates, the QC challenge extends beyond detecting parts to \emph{verifying that the right parts appear on the right variant}, under diverse appearances and operating constraints. Single-camera inspection pipelines, common in legacy deployments, struggle with occlusions, view-dependent features (e.g., roof accessories), and the throughput targets of contemporary factories~\cite{chouchene2020artificial,barman2024advancements}. In contrast, multi-camera perception can supply the complete, view-specific evidence required for reliable, real-time validation.

Prior work spans vision-based inspection systems~\cite{barman2024advancements,czimmermann2021autonomous} and quality-control frameworks~\cite{lins2025vision}, often treating inspection and defect detection as separate problems~\cite{cronin2019state,kaiser2022concept,zhou2024trvlr,zhou2019automatic,krummenacher2017wheel,li2022surface,jha2023detection}. Some approaches leverage VIN-linked databases for variant checks~\cite{wu2022vins,rak2021digital}. We adopt the practical VIN-to-manifest paradigm yet remove dependence on external lookups during perception by deriving the expected feature set up front and validating it purely from vision, which is critical when metadata access is unreliable on the line.

Three challenges remain open for Autonomous Vehicle Inspection (AVI). First, information overhead: multi-view, multi-task perception must meet real-time latency and footprint constraints~\cite{kuutti2020survey}. Second, semantic ambiguity: classical detectors answer “what is present,” while factories need “is this \emph{correct} for this variant?”, a perception-to-reasoning gap insufficiently explored in prior works~\cite{jha2023detection}. Third, defect data scarcity: rare, long-tail scratches and dents limit supervised training, motivating data-efficient solutions for industrial QA~\cite{yorulmucs2021predictive,araujo2023testing}.

\noindent We propose a deployable \emph{multi-view} AVI system that closes the perception–reasoning gap for variant-aware QC:
\begin{itemize}
    \item Targeted view assignment: A camera-to-task routing policy that restricts each detector to informative views (e.g., T2/T3 for roof features; L1/R1 for wheels), reducing false positives and latency while preserving accuracy.
    \item Semantic validation layer: A decision-level fusion and rule engine that integrates outputs from detection, classification, and OCR and compares them against a VIN-derived feature manifest to emit interpretable pass/fail decisions.
    \item Data-efficient defect module: A damage segmentation branch for scratches/dents tailored to limited labels, enabling reliable anomaly localization on side panels and bumpers.
    \item Comprehensive benchmarking: A real-line evaluation across four models with ablations (single-view, no-fusion, no-segmentation) and comparisons against recent detectors/segmenters, establishing best-per-task choices for industrial deployment.
\end{itemize}
This work demonstrates that principled multi-camera routing, combined with semantic reasoning, enables robust, explainable, and real-time variant verification in automotive manufacturing, advancing the practical state of autonomous inspection.

%% file: 02_related_work.tex
\section{Related Work}
\enlargethispage{-\baselineskip}

\begin{figure*}[h!]
    \centering
    \includegraphics[width=\textwidth, height= 15cm]{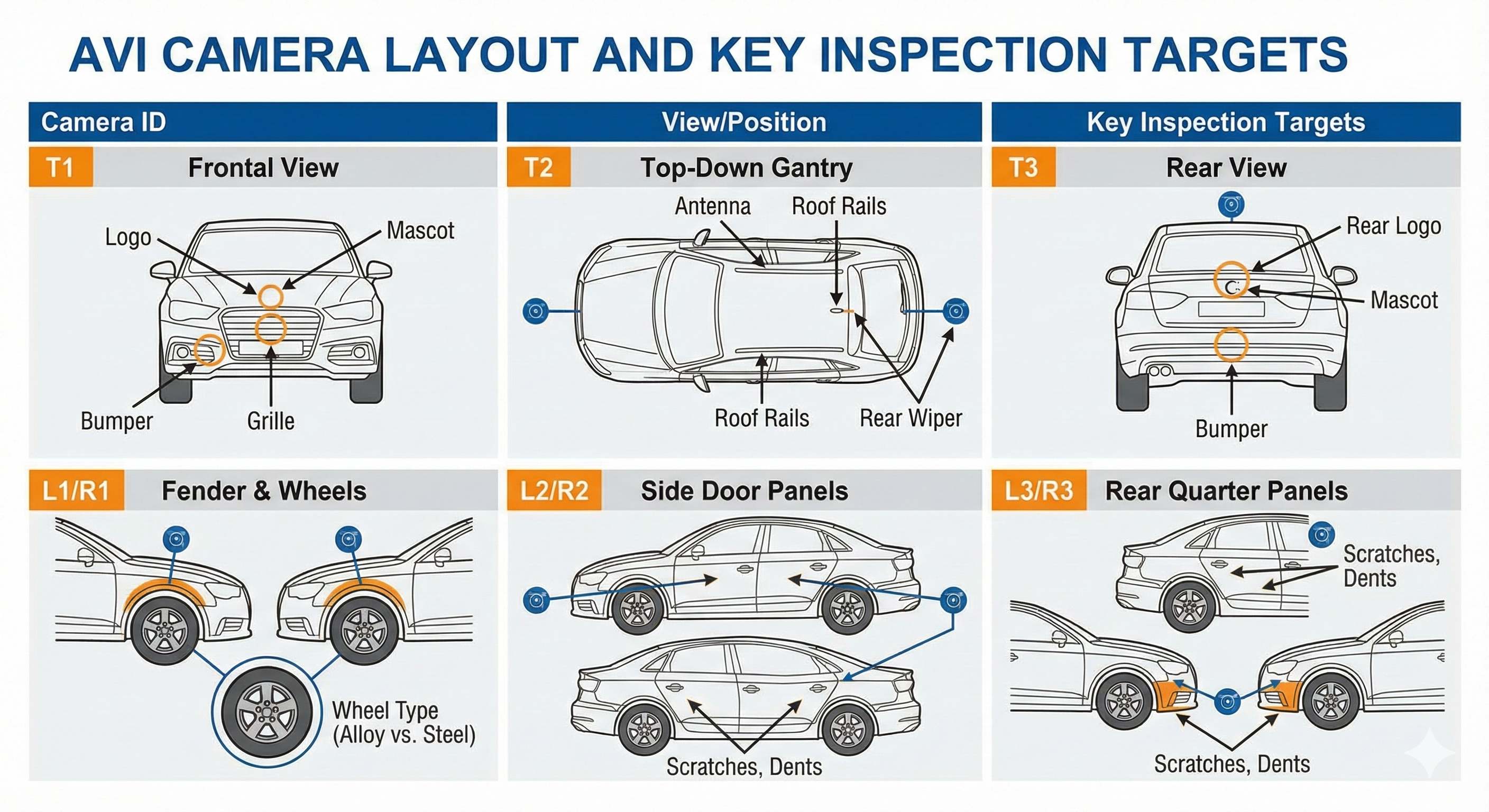}
    \caption{The Automated Vehicle Inspection (AVI) camera setup, and synchronization. The synchronized images from 11 cameras are routed to specialized perception modules to perform each specific task.}
    \label{fig:system_architecture}
\end{figure*}

\subsection{Multi-camera perception in autonomous vehicles}
Multi-camera rigs are standard in autonomous driving and robotics because they expand the field of view, add redundancy, and improve robustness in cluttered, dynamic scenes. Prior work spans camera-only and multi‑modal pipelines for vehicle detection~\cite{jha2023detection,huang2021bevdet, zhou2024trvlr}, semantic segmentation~\cite{ha2017mfnet,zhang2023cmx}, and enhancement under adverse illumination~\cite{jha2025rt,cao2023deep,wang2025thermal}. These efforts consistently show that fusing complementary viewpoints/modalities mitigates occlusion and lighting variance, properties directly useful for inspection where fine parts (e.g., rails, antennas) and glossy surfaces challenge single‑view systems. Our system adopts this principle but tailors it to factory constraints via task‑to‑view assignment and decision‑level fusion for variant validation.

\subsection{Vision‑based quality inspection and control}
Automated Optical Inspection (AOI) is long established in electronics~\cite{lee2021machine,shi2024enhancing}, yet its transfer to automotive autonomy remains partial. Existing studies address vehicle inspection~\cite{barman2024advancements,czimmermann2021autonomous} and quality control~\cite{lins2025vision} largely in isolation, limiting end‑to‑end guarantees. The literature suggests vision can raise throughput and consistency, but a unified framework that couples part verification with defect analysis is needed for modern lines. We position our work as such a unification: a single pipeline that detects variant‑specific parts and surface anomalies and renders an interpretable pass/fail decision.

\label{subsec:perception_modules}
\begin{figure*}[h]
    \centering
    \includegraphics[width=\textwidth, height= 12cm]{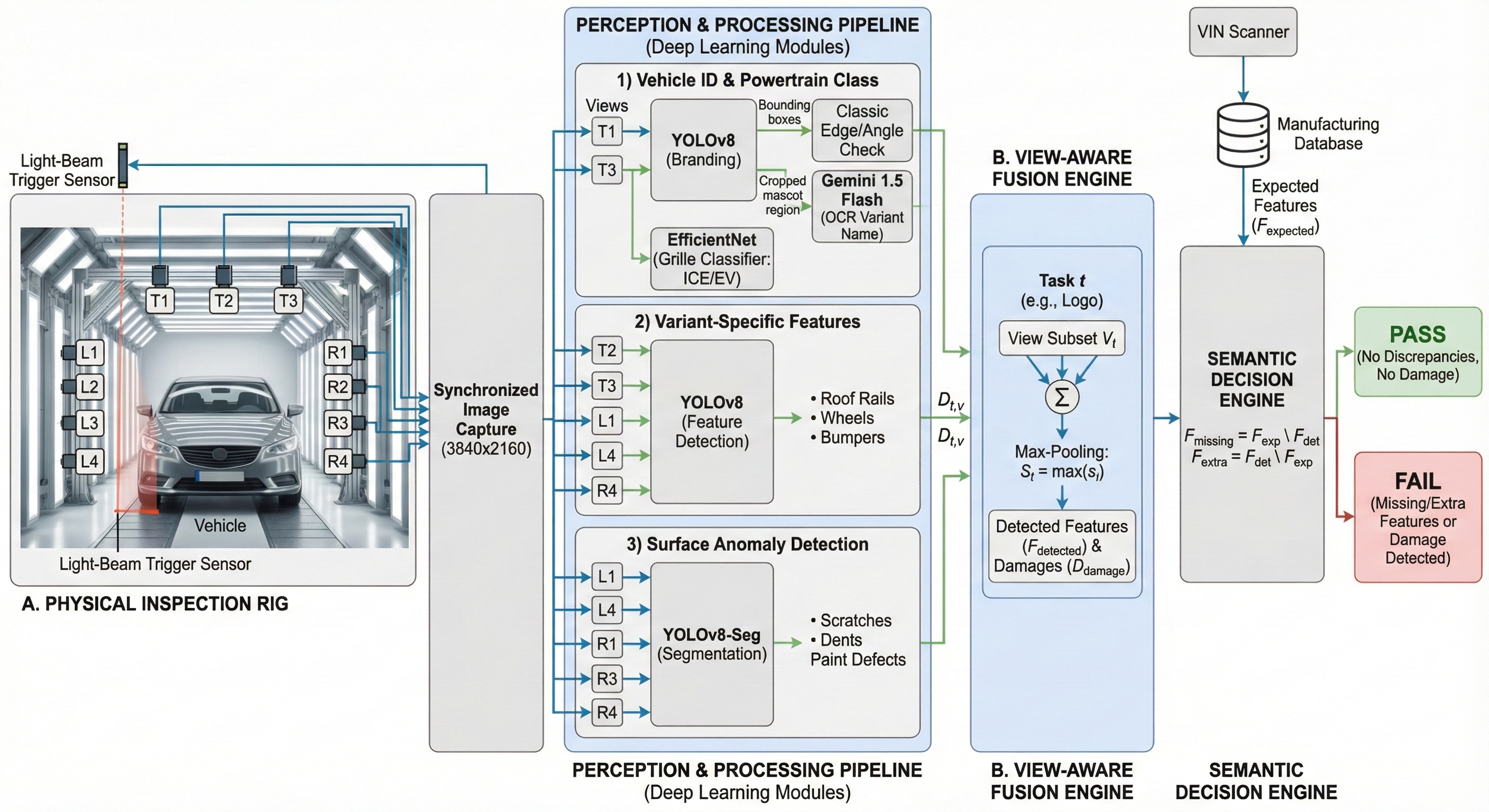}
    \caption{This Automated Vehicle Inspection (AVI) system uses an 11-camera array and deep learning to perform a comprehensive, multi-angle scan of each vehicle. The system automatically identifies the vehicle, detects features and anomalies, and cross-references data to generate an instant pass/fail report.}
    \label{fig:system_architecture}
\end{figure*}

\subsection{Information fusion and defect detection}
Classical fusion in automotive inspection often combines LiDAR and cameras at the raw/data level~\cite{zhao20223d,zhao2020fusion}, which is powerful yet sensitive to calibration and noise. Decision‑level fusion, aggregating per‑sensor inferences, offers lighter, noise‑tolerant deployment~\cite{wei2020decision, jha2025adaptive}. We follow this latter paradigm: view‑aware detectors produce standardized evidence that a semantic rule engine evaluates against a VIN‑conditioned manifest. For defects, real vehicles exhibit long‑tail, rare patterns where large labeled datasets are scarce~\cite{li2022surface,zhou2019automatic}. Detectors are therefore appropriate; in our pipeline, a segmentation module provides data-efficient damage cues that seamlessly integrate with the decision layer.

%% file: 03_method.tex

\section{Proposed Method: The AVI System}
\label{sec:method}

\begin{table}[h!]
    \centering
    \caption{AVI Camera Layout and Key Inspection Targets.}
    \label{tab:camera_layout}
    \begin{tabular}{@{}lll@{}}
        \toprule
        \textbf{Camera ID} & \textbf{View/Position} & \textbf{Key Inspection Targets} \\
        \midrule
        T1 & Frontal View & Logo, Mascot, Bumper, Grille \\
        T2 & Top-Down Gantry & Antenna, Roof Rails, Rear Wiper \\
        T3 & Rear View & Rear Logo, Mascot, Bumper \\
        L1/R1 & Fender \& Wheels & Wheel Type (Alloy vs. Steel) \\
        L2/R2 & Side Door Panels & Scratches, Dents \\
        L3/R3 & Rear Quarter Panels & Scratches, Dents \\
        L4/R4 & Side Bumpers & Scratches, Dents \\
        \bottomrule
    \end{tabular}
\end{table}

Our proposed system, the Automated Vehicle Inspection (AVI) platform, is an end-to-end multi-camera perception solution designed for autonomous quality control in automotive manufacturing. The system's novelty lies not in a single new model, but in the intelligent integration of specialized deep learning modules, a targeted view-assignment strategy, and a semantic rule engine that enables high-level, variant-aware reasoning.

\subsection{System Architecture and Camera Layout}
\label{subsec:architecture}

The AVI system is built around a fixed hardware rig consisting of 11 synchronized cameras strategically positioned along an inspection lane. A light-beam sensor triggers simultaneous image capture, ensuring a complete and coherent snapshot of the vehicle. The resolution of the camera is 3840x2160. The camera layout is optimized to provide comprehensive coverage, with each camera assigned a specific set of observational goals. A diagram of the system architecture is shown in Figure~\ref{fig:system_architecture}.

The camera configuration and its primary inspection targets are detailed in Table~\ref{tab:camera_layout}.

The AVI system employs a suite of specialized, pre-trained deep learning models, each optimized for a specific perception task. This multi-task approach ensures high performance for each sub-problem.

\begin{enumerate}
    \item Vehicle Identification \& Powertrain Classification: The system first identifies the core attributes of the vehicle. A YOLOv8  \cite{varghese2024yolov8}model trained on views T1 and T3 detects branding elements (logo, mascot). The detected mascot region is then passed to Gemini 1.5 Flash \cite{team2024gemini} for OCR to extract the variant name. Concurrently, a classic edge detection and angle check algorithm validates the rotational alignment of the logo, while an EfficientNet \cite{tan2019efficientnet} classifier uses the front grille image from camera T1 to distinguish between Internal Combustion Engine (ICE) and Electric Vehicle (EV) models.

    \item Variant-Specific Feature Detection: A second YOLOv8 \cite{varghese2024yolov8} model is tasked with identifying key configurable features. It is trained on specific views to ensure accuracy: T2/T3 for roof-level items (such as antennas and rails), and L1/R1/L4/R4 for side features (including wheel type and bumper style).

    \item Surface Anomaly Detection: To identify manufacturing defects, we use a YOLOv8-Seg model. This segmentation model is applied to all side-facing camera views (L1-L4, R1-R4) to detect and generate pixel-level masks for scratches, dents, and paint deformities. 
\end{enumerate}

\subsection{View-Aware Fusion and Semantic Decision Engine}
\label{subsec:fusion_logic}

The core novelty of the AVI system lies in how it fuses the outputs of these perception modules into a final, coherent decision. Let $\mathcal{T}$ be the set of all inspection tasks (e.g., `logo`, `antenna`). For each task $t \in \mathcal{T}$, we define a subset of camera views $V_t \subseteq \{T1, \dots, R4\}$ that are assigned to it.

\paragraph{View-Aware Fusion.}
For a given task $t$, each assigned camera view $v \in V_t$ produces a set of detections $D_{t,v} = \{(b_i, s_i)\}$, where $b_i$ is a bounding box and $s_i$ is its confidence score. The fusion process aggregates this evidence to produce a single detection score for the task, $S_t$. We use a max-pooling strategy over all views for robustness:
\begin{equation}
S_t = \max_{v \in V_t, (b_i, s_i) \in D_{t,v}} (s_i)
\end{equation}
A feature is considered present if its fused score $S_t$ exceeds a task-specific threshold $\tau_t$. The final set of detected features is thus defined as:
\begin{equation}
\mathcal{F}_{\text{detected}} = \{t \in \mathcal{T} \mid S_t > \tau_t\}
\end{equation}


\paragraph{Semantic Decision Engine.}
Upon scanning a vehicle's VIN, a set of expected features, $\mathcal{F}_{\text{expected}}$, is retrieved from a database. The decision engine then performs a set-theoretic comparison to identify discrepancies. The sets of missing and extraneous features are calculated as:
\begin{align}
\mathcal{F}_{\text{missing}} &= \mathcal{F}_{\text{expected}} \setminus \mathcal{F}_{\text{detected}} \\
\mathcal{F}_{\text{extra}} &= \mathcal{F}_{\text{detected}} \setminus \mathcal{F}_{\text{expected}}
\end{align}
The final inspection result is determined by the contents of these sets, along with the set of detected damages, $\mathcal{D}_{\text{damage}}$. A "PASS" is issued only if both the discrepancy sets and the damage set are empty.

%% file: 04_experiment.tex
\section{Experiments and Results}
\label{sec:experiments}
We evaluate both the individual perception modules and the integrated AVI pipeline to answer: (1) which off‑the‑shelf backbones are best per task; (2) how much view‑aware fusion and the semantic rule engine contribute to end‑to‑end accuracy; and (3) whether the system meets factory throughput constraints.

\subsection{Dataset}
\label{subsec:dataset}
We construct a hybrid dataset spanning private, real‑line imagery and public resources. The size of the dataset is more than 25000 images:

\begin{figure*}[h!]
    \centering
    \includegraphics[width=\textwidth, height = 15cm]{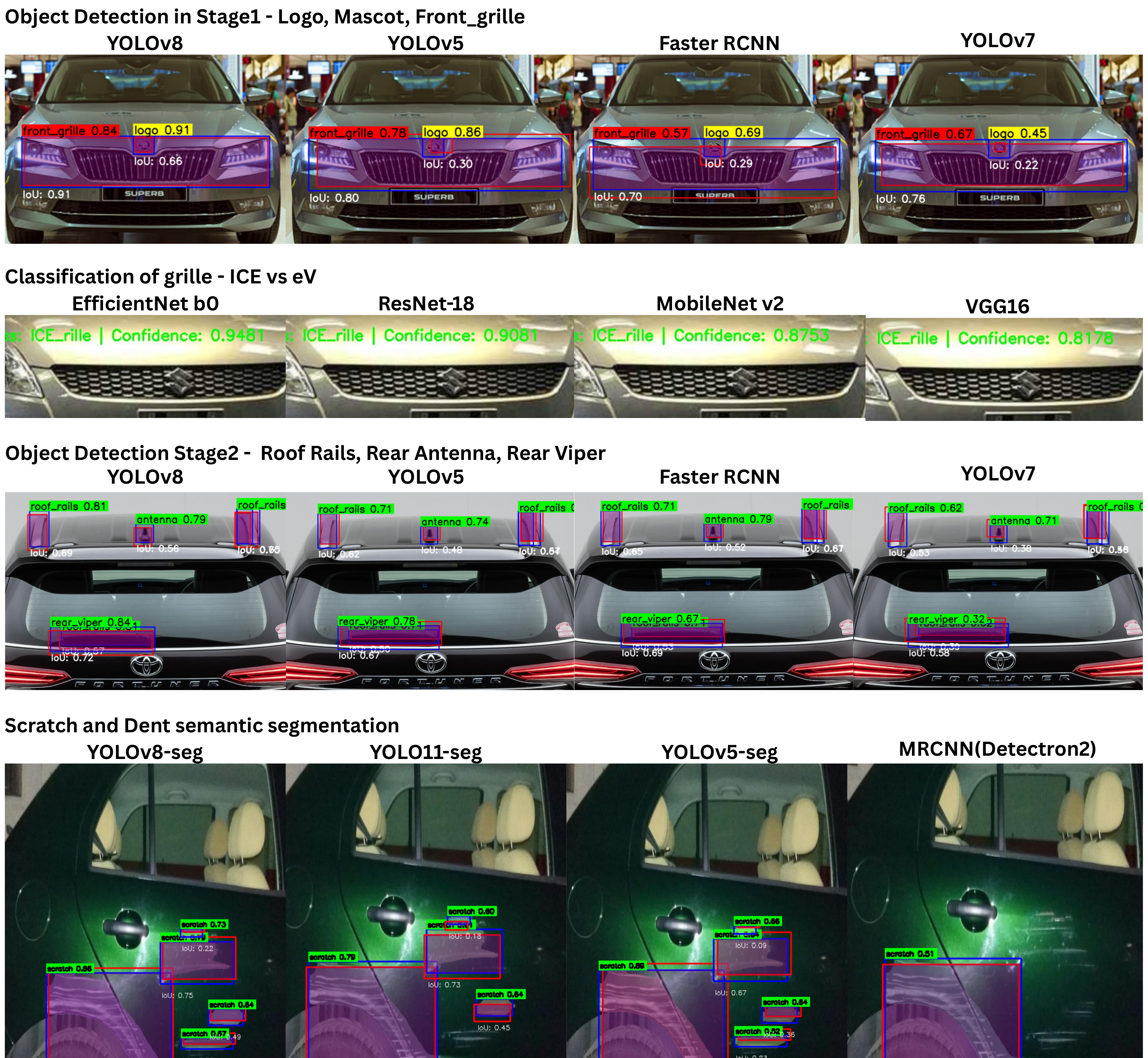}
    \caption{Qualitative results. Top: Stage~1 branding detection (logo, mascot, front\_grille) across backbones. Second: grille crops for ICE/EV classification. Third: Stage~2 variant features (roof rails, rear antenna, rear wiper, wheel type). Bottom: scratch/dent instance segmentation masks.}
    \label{fig:qualitative}
\end{figure*}

\begin{enumerate}
    \item OEM Vehicle Dataset (Private). 
    High-resolution, multi-view image captures obtained from an automotive OEM production environment, covering four distinct vehicle models and multiple variants. Each sample is paired with a manifest derived from vehicle identification metadata, specifying expected attributes (e.g., wheel type, roof rails, antenna, rear wiper).
    \item Roboflow Universe. Publicly available datasets from the Roboflow Universe were employed, providing annotated images of vehicle parts and surface defects, including scratches and dents, hence complementing the OEM dataset for training and evaluation of the system.
\end{enumerate}

We split the data into \(\,80/10/10\,\) train/val/test, ensuring vehicle‑level disjointness across splits to prevent leakage.

\subsection{Tasks \& Evaluation Metrics}
\label{subsec:metrics}
We report component‑level, system‑level, and operational metrics.

\paragraph{Component metrics.}
\begin{itemize}
    \item Object detection (parts). Precision, Recall, and mAP@50 (IoU threshold \(=0.5\)).
    \item Damage segmentation. mIoU for pixel‑level masks (scratches, dents).
    \item Binary classification (ICE vs. EV). Accuracy and F1‑Score.
\end{itemize}

\paragraph{System‑level verification accuracy.}
The end‑to‑end objective is to match the final \emph{Pass/Fail} verdict to the ground truth manifest:
\[
Acc_{\text{sys}}=\frac{1}{N}\sum_{i=1}^{N}\mathbb{I}\!\left(\text{Verdict}_i=\text{GroundTruth}_i\right),
\]
where \(N\) is the number of vehicles and \(\mathbb{I}(\cdot)\) is the indicator function.

\subsection{Qualitative Analysis}
\label{subsec:qualitative}
Figure~\ref{fig:qualitative} shows representative outputs across four stages, highlighting correct detections and evidence clarity used by the rule engine, and all these images are from an open-sourced test dataset.

\noindent Stage 1 (Branding). YOLOv8 yields tight, low‑overlap boxes for \emph{logo}, \emph{mascot}, and \emph{front\_grille}; alternative backbones show occasional misses or oversized boxes, consistent with the precision/recall ranking and improved OCR and grille‑cue stability.\\
ICE/EV cues. Grille crops retain texture/aperture patterns; EfficientNet/ResNet produce confident logits even with specular highlights and partial occlusions.\\
Stage 2 (Variant features) From top/rear views, YOLOv8 separates \emph{roof rails}, \emph{rear antenna}, and \emph{rear wiper} from roof edges and low‑contrast stubs; side views provide crisp wheel‑type evidence for manifest checks.\\
Stage 3: (Damage masks) YOLOv8‑Seg returns compact, high‑IoU masks on fine scratches and shallow dents; competing models under‑ or over‑segment reflective regions, matching the mIoU ranking.

\subsection{Quantitative Results}
\label{subsec:quantitative_results}
We first benchmark backbones per task, then evaluate system‑level performance.

\subsubsection{Stage 1: Vehicle Identification and Branding Detection}
We detect \emph{logo}, \emph{mascot}, and \emph{front\_grille} as inputs to OCR and ICE/EV classification. Table~\ref{tab:stage1_detection_results} reports Precision(P), Recall(R), Mean Average Precision(mAP)@50, and F1.

\begin{table}[h!]
\centering
\caption{Stage~1: object detection for branding cues.}
\label{tab:stage1_detection_results}
\begin{tabular}{lcccc}
\toprule
Model & Precision & Recall & mAP@50 & F1 \\
\midrule
Faster R\textendash CNN~\cite{ren2015faster} & 0.8356 & 0.7925 & 0.8548 & 0.8135 \\
YOLOv5 & 0.8727 & 0.7727 & 0.9034 & 0.8196 \\
YOLOv7~\cite{wang2023yolov7} & 0.5438 & 0.4149 & 0.5991 & 0.4707 \\
\textbf{YOLOv8}~\cite{varghese2024yolov8} & \textbf{0.9264} & \textbf{0.8542} & \textbf{0.9572} & \textbf{0.8889} \\
\bottomrule
\end{tabular}
\end{table}

\subsubsection{Stage 2: Variant‑Specific Feature Detection}
We detect \emph{roof rails}, \emph{rear antenna}, and \emph{rear wiper} from top/rear views and side cues for wheel type. Results in Table~\ref{tab:stage2_feature_detection_results} again favor YOLOv8.

\begin{table}[h!]
\centering
\caption{Stage~2: variant feature detection.}
\label{tab:stage2_feature_detection_results}
\begin{tabular}{lcccc}
\toprule
Model & Precision & Recall & mAP@50 & F1 \\
\midrule
Faster R\textendash CNN~\cite{ren2015faster} & 0.8356 & 0.7925 & 0.8548 & 0.8135 \\
YOLOv5 & 0.8727 & 0.7727 & 0.9034 & 0.8196 \\
YOLOv7~\cite{wang2023yolov7} & 0.5438 & 0.4149 & 0.5991 & 0.4707 \\
\textbf{YOLOv8}~\cite{varghese2024yolov8} & \textbf{0.9264} & \textbf{0.8542} & \textbf{0.9572} & \textbf{0.8889} \\
\bottomrule
\end{tabular}
\end{table}

\subsubsection{Stage 3: Surface Anomaly Segmentation}
We compare instance‑segmentation backbones for scratches/dents (Table~\ref{tab:stage3_segmentation_results}); YOLOv8‑Seg attains the best mAP and mIoU. The BB stands for bounding box, in (Table~\ref{tab:stage3_segmentation_results}).

\begin{table}
\centering
\caption{Stage~3: scratch/dent segmentation.}
\label{tab:stage3_segmentation_results}
\begin{tabular}{lcccccc}
\toprule
Model & P\,(BB) & R\,(BB) & P\,(Mask) & R\,(Mask) & mAP & mIoU \\
\midrule
Mask R\textendash CNN & 0.548 & 0.501 & 0.516 & 0.498 & 0.463 & 0.443 \\
YOLOv5\textendash seg & 0.764 & 0.705 & 0.760 & 0.689 & 0.744 & 0.732 \\
YOLO11\textendash seg & 0.841 & 0.763 & 0.838 & 0.757 & 0.819 & 0.803 \\
\textbf{YOLOv8\textendash seg} & \textbf{0.856} & \textbf{0.858} & \textbf{0.848} & \textbf{0.853} & \textbf{0.890} & \textbf{0.879} \\
\bottomrule
\end{tabular}
\end{table}

\clearpage
\subsection{Ablation Study}
\label{subsec:ablation}
We quantify the impact of multi‑view design and the damage module (Table~\ref{tab:ablation_study}). Single‑view baselines lack feature coverage; removing segmentation reduces defect recall despite lower latency. Here, Feat. Cov. (\%) denotes Feature Coverage, Def. Det. (\%) represents Defect Detection, Lat. (ms) refers to Latency, and Verif. Acc. (\%) corresponds to Verification Accuracy. The multi‑view rig is necessary for complete coverage; the segmentation branch raises defect detection by \(\approx\)14–18 pp with an acceptable latency trade‑off, yielding a balanced system suitable for deployment.

\begin{table}[!t]
\centering
\caption{Ablations: multi‑view value and segmentation contribution.}
\label{tab:ablation_study}
\begin{tabular}{lcccc}
\toprule
\textbf{Case} & \textbf{Feat. Cov. (\%)} & \textbf{Def. Det. (\%)} & \textbf{Lat (ms)} & \textbf{VA (\%)}\\
\midrule
T1 (Front) & \(2/7 \rightarrow \approx 28.6\) & 35–45 & 140–155 & 88 \\
T2 (Top) & \(3/7 \rightarrow \approx 42.9\) & 25–35 & 140–155 & 92 \\
T3 (Rear) & \(5/7 \rightarrow \approx 71.4\) & 45–60 & 150–170 & 93 \\
Side & \(1/7 \rightarrow \approx 14.3\) & 20–30 & 140–155 & 91 \\
\midrule
No Seg. (all views) & 96.1 & 68–72 & 215 & 97 \\
\textbf{Full AVI System} & \textbf{98.7} & \textbf{86.5} & \textbf{285} & \textbf{93} \\
\bottomrule
\end{tabular}
\end{table}

%% file: 05_discussion.tex
\section{Discussion}

\subsection{Bridging the Perception-Reasoning Gap in Industrial Autonomy}
While recent advances in deep learning have solved the what is it problem (detection), industrial autonomy requires answering the is it correct problem (verification). Standard object detectors lack the contextual awareness to distinguish between a valid missing part (e.g., a base-model vehicle lacking a sunroof) and a defect (e.g., a premium model missing a sunroof). Our work demonstrates that relying solely on data-driven perception is insufficient for variant-aware inspection. The AVI system bridges this gap through its Semantic Decision Engine. By decoupling visual detection from logic to validating standardized evidence against a VIN-derived configuration, we transform probabilistic detections into deterministic pass/fail decisions. This architecture suggests a blueprint for future autonomous quality control systems: perception modules should act as sensors that feed a deterministic reasoning layer, rather than attempting to learn complex business logic end-to-end. 

\subsection{The Necessity of Multi-View Fusion for Robustness}
A critical critique of autonomous visual inspection is the reliance on single-viewpoint sensors, which fail under occlusion. Our ablation study (Table~\ref{tab:ablation_study}) empirically quantifies this feature coverage risk. Single views (Front T1 or Rear T3) achieve only 28-71\% feature coverage, leading to unacceptable verification accuracy (88-93\%). The No Segmentation baseline highlights another trade-off: while excluding the segmentation module reduces latency by $\approx$70 ms, it causes a catastrophic drop in defect detection (from 86.5\% to $\approx$70\%)5. This confirms that in autonomous industrial settings, sensor redundancy (11 cameras) and task specialization (segmentation vs. detection) are not inefficiencies, but rigid requirements for reliability. 

\subsection{Deployability vs. Complexity in Industry 4.0}
Unlike academic benchmarks that prioritize mAP improvements, real-world deployment is constrained by the tact time of production lines. The AVI system sustains a throughput of 3.3 vehicles/minute ($\approx$18 seconds/vehicle) with a processing latency of 285 ms. This performance is achieved by avoiding heavy, monolithic transformers in favor of optimized, task-specific backbones (YOLOv8, EfficientNet). While newer architectures may offer marginal accuracy gains, our results show that the primary bottleneck in industrial autonomy is not detector precision, but the integration of multi-modal evidence (OCR, classification, segmentation) into a unified decision. The choice of established backbones ensures stability and ease of retraining, a critical factor for factory adoption where data for new defects is often scarce. 

\subsection{Limitations and Future Directions}
While our View-Aware Fusion layer effectively standardizes evidence, it currently relies on a static camera-to-task mapping. Future iterations could explore adaptive view attention, where the system dynamically queries different cameras based on confidence scores or detected ambiguities (e.g., zooming in on a scratch if the L1 view is occluded). Additionally, while the VIN-conditioned logic handles known variants, extending the system to open-world anomaly detection, identifying completely unknown foreign objects without prior training, remains a frontier for autonomous inspection research.

%% file: 06_conclusion.tex
\section{Conclusion}
We introduced AVI, a deployable multi-camera perception system that performs variant-aware vehicle inspection and defect detection in a single, unified pipeline. By (i) assigning tasks to views, (ii) fusing detections through a lightweight max-pool scheme, and (iii) enforcing variant rules via a semantic engine, AVI achieves complete feature coverage with industry-grade throughput. Extensive quantitative studies show that YOLOv8 outperforms earlier detectors for both branding and variant features, while YOLOv8-Seg delivers state-of-the-art scratch/dent masks (mIoU $=0.879$). An ablation analysis confirms the necessity of the full 11-camera rig and the dedicated segmentation module: removing either degrades verification accuracy or defect recall by more than 10 pp.